\documentclass{article}

\usepackage[dblblindworkshop, final]{neurips_2026}
\workshoptitle{Child Safety in AI}

\setcitestyle{authoryear,round}
\usepackage[utf8]{inputenc} 
\usepackage[T1]{fontenc}    
\usepackage{hyperref}       
\usepackage{url}            
\usepackage{booktabs}       
\usepackage{amsfonts}       
\usepackage{nicefrac}       
\usepackage{microtype}      
\usepackage{xcolor}         

\usepackage{amsmath}
\usepackage{graphicx}
\usepackage{booktabs}
\usepackage[size=tiny]{todonotes}
\usepackage{makecell}

\title{AYO: A Safe and Privacy-Preserving Platform for Studying Children’s Interactions with LLMs}

\author{
  Huiwen Alex Yang \\
  Department of Psychology \\
  University of California, Berkeley \\
  \texttt{hw.yang@berkeley.edu}
  \And
  Rachel Skwersky \\
  Department of Psychology \\
  University of California, Berkeley \\
  \texttt{rskwersky2025@berkeley.edu}
  \And
  Bill D. Thompson\thanks{These authors contributed equally.} \\
  Department of Psychology \\
  University of California, Berkeley \\
  \texttt{wdt@berkeley.edu}
  \And
  Celeste Kidd\footnotemark[1] \\
  Department of Psychology \\
  University of California, Berkeley \\
  \texttt{celestekidd@berkeley.edu}
}

\begin{document}
\maketitle
\begin{abstract}
  Children and adolescents are increasingly interacting with generative AI and large language models (LLMs), yet little is known about how these interactions may influence their development, beliefs, and behavior. Existing research has relied heavily on surveys, while direct study of children's interactions with LLMs presents substantial challenges related to safety, privacy, consent, and experimental control. We introduce \textbf{AYO} (AI Youth Open Science), an open-source platform designed for safe and systematic research on children’s interactions with conversational AI. AYO provides a child-friendly interface, configurable age-based safety guardrails, researcher monitoring, and privacy-preserving data practices. These capabilities enable cross-sectional, longitudinal, and experimental research on how children use and respond to LLMs and how AI design may shape trust, epistemic judgment, curiosity, and other aspects of development. AYO provides the infrastructure to generate evidence that guides safer AI systems, educational practices, and policies for children growing up with AI.
\end{abstract}

\section{Motivation \& Related Work}

\subsection{Children's Interactions with Conversational AI}


Generative AI and large language models (LLMs) are an increasingly common part of young people's digital environments. In 2025, a nationally representative survey of U.S. teens ages 13–17 reported 64\% of participants have used AI chatbots, including approximately three in ten who reported using them daily \citep{faverioTeensSocialMediaa, ComprehensiveReportTeens}. Similar patterns have been reported internationally for children and adolescents \citep{staksrudEuropeanChildrensUse2026, TalkingMachinesChildrens, ChildrenYoungPeople}. Frequent interactions with AI systems have the potential to impact children's learning, relationships, character development, and other aspects of their well-being and beliefs in ways that are currently not well understood.


\subsection{Challenges in Studying Children's AI Use}


Much existing research on children's and adolescents' use of AI has relied on survey-based methods \citep{faverioTeensSocialMediaa, mcclainHowTeensUse2026}. Surveys provide valuable information about the prevalence and reported purposes of AI use, but offer limited insight into how young people actually interact with AI systems. Importantly, young people's interactions with AI systems extend beyond narrowly educational contexts. Adolescents report using generative AI for entertainment, personal advice, and other forms of everyday interaction, suggesting that LLMs increasingly function not only as task-completion tools but also as information sources, guides, and social companions. These uses raise emerging concerns about how young people form relationships with AI systems, including the use of chatbots for companionship, and how repeated interactions with AI may influence cognitive and social development \citep{drugaHeyGoogleIt2017, sunAICompanionsAdolescent2026}. However, understanding these effects requires methods that can assess children’s behavior as they engage directly with AI systems, rather than relying on surveys. Considerably less is known about how younger children interact with AI systems. Existing surveys have focused primarily on adolescents, leaving an additional important gap in our understanding of how younger children navigate open-ended interactions with LLMs. 


Studying interactions between children and AI systems presents a range of methodological, ethical, and practical challenges, including recruiting and enrolling families, obtaining consent, protecting sensitive conversation data, and ensuring children's safety during open-ended interactions. Unlike laboratory experiments in which researchers can specify and control the stimuli presented to children, interactions with LLMs are open-ended. This creates challenges for safety: research found that nearly half of adolescent chatbot users in the U.S. reported at least one harmful or unsafe experience, including exposure to false information, manipulation, risky-behavior prompts, and self-harm or suicidal messages \citep{hindujaRisksHarmsConversational2026}. These concerns are particularly salient given evidence that children are increasingly using chatbots for advice and emotional support, contexts in which inaccurate or harmful responses may have consequences beyond a single informational query \citep{GenerativeAIRisks}. At the same time, conversational interactions can involve highly sensitive information, including children’s interests, moods, fears, and other personal details, creating additional challenges for privacy and data governance \citep{GuidanceAIChildren2025}. Thus, studying children’s interactions with AI systems goes beyond providing access to a chatbot: research environments must incorporate safeguards for safety, privacy, and monitoring while preserving the open-ended interactions that researchers seek to study.

\subsection{Motivation for AYO}


The challenges described above highlight a gap in the infrastructure available for studying children’s interactions with LLMs. Existing approaches often require researchers to choose between tightly controlled experimental tasks, which may not capture the open-ended nature of children’s interactions with AI, and general-purpose commercial AI systems, which may offer limited control over the safety, privacy, and data-collection requirements of research with children. 


To address this gap, we developed \textbf{AYO} (AI Youth Open Science), an open-source platform designed specifically for research on children’s interactions with LLMs. AYO is designed around the specific developmental, safety, and privacy needs of child participants. First, AYO provides a child-friendly interface and interaction design intended to make LLM interactions accessible and appropriate for children. Second, AYO incorporates multiple layers of safety mechanisms, including automated programmatic guardrails as well as human monitoring and oversight, to identify and respond to potentially harmful content and help maintain a safe interaction environment. Finally, AYO is designed around privacy-preserving research practices, including a zero-data-retention policy, to minimize the collection and storage of sensitive information generated during children’s interactions with LLMs. Together, these features provide researchers with a controlled and privacy-preserving environment for studying open-ended child–LLM interactions while maintaining safeguards designed to protect children from harmful content and interactions.

\section{AYO Architecture}
\subsection{Recruitment}


Studying children’s interactions with commercial AI systems poses unique privacy and data retention challenges, as these model providers typically have policies that restrict the use of their products by children under the age of 13. For example, OpenAI previously prohibited the use of their ChatGPT model by those under 18 without parental consent. As of August 18, 2026, OpenAI has deployed ChatGPT for Teens for users that their system classifies as 13 to 17 years old \citep{IntroducingChatGPTTeens2026}. 
AYO addresses this constraint by allowing researchers to deploy study-specific models through researcher-controlled infrastructure. Researchers can use open-weight models hosted on their own infrastructure or, where appropriate, access commercial models through research-specific agreements governing data retention and safety policies.


AYO's infrastructure for enrollment and data collection supports child-centered consent and privacy practices, including compliance with the Children’s Online Privacy Protection Rule, 16 C.F.R. § 312 (2024) \citep{ftc_coppa_amendments_2025}. The Federal Trade Commission mandates that verifiable parental consent be obtained under COPPA for all operators of online services that collect the personal information and data of users under 13 years of age \citep{ftc_coppa_amendments_2025}. For participants under 13, we will use identity-verification video calls to establish the parent-child relationship and obtain verifiable parental consent, following precedent from internet-based and social media research \citep{hokkeEthicalIssuesUsing2018}. We will retain documentation of consent but not video recordings to minimize participant identifiability. To accommodate children and teenager participants without personal email accounts or devices, parents and guardians may choose whether to use a family or participant email for account creation, with permission for its storage and use.


\subsection{Software}
\begin{figure}[htbp]
    \centering
    \includegraphics[width=0.8\textwidth]{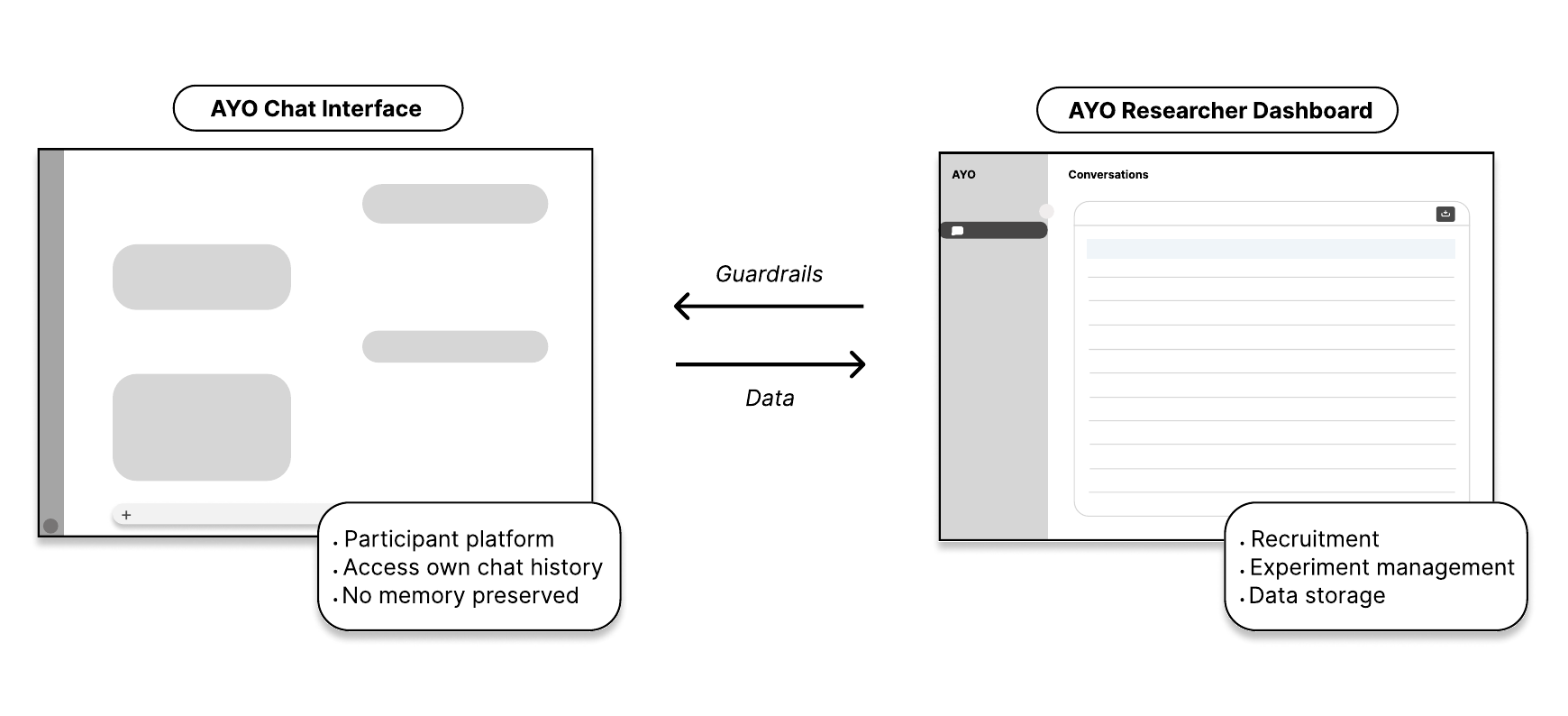}
    \caption{The AYO System Design. A participant-facing LLM chat interface and a researcher dashboard to manage participants, guardrails, and data storage.}
    \label{fig:ayo_diagram}
\end{figure}

The AYO system has two components: a participant-facing chat interface and an internal researcher dashboard. The components are built on two core open-source technologies, LibreChat, a configurable chat interface, and LiteLLM, a proxy which manages calls to an LLM API provider. AYO is fully customizable – the participant interface can be flexibly modified and the model settings can be reconfigured to align with the needs of ongoing studies. The chat interface was designed to be child-friendly while preserving the original user experience of interacting with consumer conversational products such as ChatGPT. The interface has text-to-speech functionality and file uploading enabled, but minimally exposed menu options so that participants don't have access to advanced configuration controls. Users are able to access their own chat history, but no memory is preserved across chats. The AYO dashboard serves as an experiment management platform for the storage of participant queries and model responses. Researchers can invite participants through the dashboard by sending templated emails with secure, unique sign-up links and organize participants into cohorts to manage experimental conditions. Conversation data can be viewed directly in the dashboard, or exported with relevant metadata, such as time, location, conversational turns, and generated attachments. The dashboard exposes two crucial safety mechanisms – configurable model guardrails with age based cohort parameterization and logged alerts if a guardrail is triggered. If a user triggers a guardrail, the alert is logged in the conversation database, and the user receives an appropriate redirection message or hard block depending on the conversational context.

\subsection{Data, Privacy \& Safety}

AYO’s data handling practices are designed to minimize the risk of exposing sensitive and unfiltered conversation data. To prioritize users' safety and privacy, all identifying information, including consent materials and survey instruments, is stored separately from conversation data. If a participant reveals personally identifiable information in a chat, the model will trigger a guardrail, blocking the request before it reaches the model provider. 


We are employing a multifaceted approach to ensure the safety of AYO’s users. In our ongoing studies, the data collection period will be limited to 6 weeks per participant. We systematically evaluated 22 child-safety keyword-blocking guardrails developed by LiteLLM, the open-source software acting as a middle layer between the AYO interface and the model API. We conducted thorough testing to determine the model’s behavior in response to prompts that should be allowed, prompts that should be blocked, adversarial attempts to bypass safety features, and a range of edge cases. To supplement gaps we identified in the testing process, we designed a semantic guardrail to catch harmful content and respond sensitively through an evaluator prompt. The guardrail is expansive and context-based, covering categories such as romantic content or persona adoption, self-harm intent, bias or stereotyping, graphic violence, and sexual content.

Given the rising trend of adolescents interacting with large language models for emotional support or medical questions \citep{oneil2026census}, studies of this kind require a protocol for notifying families if their child is using the platform to discuss self-harm, suicidality, and emotional distress. We devised a crisis escalation protocol to categorize harmful interactions with the system into tiers and define the appropriate response for communicating with guardians in a timely manner. Trained research staff manually review platform activity and maintain a coverage log of any system-flagged incidents. Content can be classified into low, medium, or high severity tiers with corresponding response procedures, ranging from continued monitoring, parental notification, platform access suspension, and mandatory reporting as required by law. High severity incidents trigger outreach directly by phone in addition to email, and provision of crisis resources to families.

\section{Applications}


AYO enables research on how children interact with and are influenced by conversational AI, supporting both cross-sectional and longitudinal studies. Researchers can examine how children’s interactions with LLMs vary across developmental stages, how children evaluate and respond to information provided by LLMs, and how they develop beliefs about AI systems. AYO can also be used to study how children use LLMs for different purposes, including information-seeking, problem-solving, entertainment, and social interaction.

AYO gives researchers fine-grained control over the interaction environment. Researchers can configure safety guardrails based on participants’ age, research questions, and desired degree of experimental control, as well as define interaction parameters and study-specific interventions. In addition, AYO also supports near-real-time collection of participant feedback. For example, researchers can automatically administer a questionnaire immediately following a child’s interaction with the platform, capturing the child’s perceptions and experiences. This combination of configurable interaction environments and immediate behavioral measures enables researchers to investigate how different LLM behaviors and designs influence children’s trust, epistemic judgment, and other aspects of development as they interact with a conversational AI. AYO also provides an infrastructure for investigating whether LLMs can be designed to support positive developmental outcomes, including character development, curiosity, and belief formation, while minimizing potentially harmful influences. Beyond individual studies, AYO is an open-science platform built to support a broader research program at the intersection of developmental science and AI safety. Future research can generate the cross-sectional and longitudinal evidence needed to understand how AI shapes children’s development and generate evidence-based design practices and policy guidance for model developers, educators, and policymakers. AYO provides infrastructure for addressing a growing gap between children’s rapidly increasing use of AI and our scientific understanding of its consequences. As children inevitably incorporate AI into their everyday lives, researchers, policymakers, and educators need empirical evidence to understand how these systems shape children’s development and to determine how they can be governed safely. By enabling systematic study of children’s interactions with LLMs, AYO will help generate the evidence needed to guide the development of safer AI systems, science-informed policies, and educational practices for a generation growing up with AI.



\newpage
\bibliographystyle{plainnat}
\bibliography{references,additional_references}





\end{document}